\documentclass[11pt]{article}

\usepackage[T1]{fontenc}
\usepackage[utf8]{inputenc}
\usepackage{lmodern}
\usepackage[margin=1in]{geometry}
\usepackage{microtype}
\usepackage{booktabs}
\usepackage{graphicx}
\graphicspath{{figures/}}
\usepackage{placeins}
\usepackage{array}
\usepackage{enumitem}
\usepackage{hyperref}
\usepackage{xcolor}

\hypersetup{
  colorlinks=true,
  linkcolor=blue!50!black,
  urlcolor=blue!50!black,
  citecolor=blue!50!black
}

\title{The Power of Meta Access in the Software of the Future --- Demonstrated on LLM Chat UIs and Chat AI}
\author{
  Julian Kunkel\\
  Universität Göttingen, GWDG\\
  \href{mailto:julian.kunkel@gwdg.de}{julian.kunkel@gwdg.de}
  \and
  Ali Doosthosseini\\
  Universität Göttingen\\
  \href{mailto:ali.doosthosseini@uni-goettingen.de}{ali.doosthosseini@uni-goettingen.de}
}

\date{\today}

\newcommand{\metachat}{\textsc{Meta Chat}}

\begin{document}

\maketitle

\begin{abstract}
Chat remains the default surface for many large language model (LLM) tasks, even as agentic
tools such as OpenCode excel at coding and automation. Yet the model sees the current dialogue
but not the workspace around it. We explore meta access---giving the assistant a structured,
permissioned view of the interface itself and propose a ladder of levels: feature tutoring; metadata, settings,
and preference access; retrieval over summaries and messages, confirmed mutation, and full
orchestration. Meta access can turn a chat client from many screens into one entry
point for finding, navigating, organising, and operating it, while raising risks around grounding,
privacy, prompt injection, and trust. General-purpose agentic harnesses are runtimes; 
because they can change their own settings, meta access becomes reflexive
and needs a separate control-plane boundary. We ground this in the Meta Chat prototype of
the Chat AI client. It builds bounded context from, e.g., summaries, settings, and relevant
messages; links references to conversations, topics, and personas;
supports confirmed, reversible changes; offers a browsable tutorial library; and recommends
personas as user-started conversations. By exposing application state in structured form, meta
access can extend screen readers from widget descriptions to semantic workspace
operations. Meta access thus turns chat from a window onto one conversation into a control
layer for the whole workspace, extending chat toward agentic capability without surrendering
user control. As assistants become embedded in IDEs, dashboards, office suites, and operating
systems, AI-era interfaces should expose a structured, permissioned model of their state and
capabilities, ideally by opening their existing metamodels through a debugger-like interface.
\end{abstract}

\noindent\textbf{Keywords:} LLM chat interfaces, AI-era interfaces, meta access,
application metamodels, interface agents, conversational search, feature
tutoring, personas, guardrails, tool use, human--computer interaction,
accessibility, local-first software.

\section{Introduction}

The chat interface has been the default surface for interacting with large
language models, and it remains the most common one for many day-to-day tasks: drafting,
search, planning, learning, and personal assistance. In parallel, a second
interaction model has matured. Agentic tools such as OpenCode
\cite{opencode2026} operate in a terminal or editor, call tools, and carry out
multi-step tasks on external environments; they have become a natural choice
for much software work. The
two models are complementary rather than substitutes. Agents excel when the
task is to operate on an environment, while chat UIs remain the natural
surface when the user wants to ask, compare, plan, draft, or re-find
something, and when the conversation itself is the workspace.

The deeper claim of this paper is architectural rather than conversational.
Chat is the first surface on which meta access is practical, but the pattern is
not specific to chat. An emerging AI-era interface is one in which an assistant
is embedded in the application rather than confined to a separate chat window,
and in which the assistant can reason over the application's own model of its
state and capabilities. Whether the surface is a chat client, an IDE, a
spreadsheet, a dashboard, or an operating system, the same asymmetry appears:
the assistant sees the current task but not the workspace that contains it.
Meta access is the missing contract between the two. The contrast is sharpest
for general-purpose agentic harnesses such as OpenCode, which are runtimes
rather than applications with a fixed purpose, and which can change the
configuration that governs their own tools and permissions
(Section~\ref{sec:harnesses}).

In applications such as Chat AI \cite{doosthosseini2026saia}, that workspace is
larger than the current conversation: a sidebar of past conversations,
user-defined topics, model selection, attachments, memories, settings, a
persona library, and feature-specific controls. The model, however, usually
sees only the current conversation plus a system prompt. Everything that makes
the workspace a workspace---what else the user
has discussed, how their chats are organised, which features exist and where
they live---is outside the model's view.

This asymmetry creates a practical problem. As a user's history grows, the
interface becomes harder to navigate: finding a conversation about a past
decision, remembering which topic contains it, or learning a rarely used
feature requires manual scanning. Conventional search helps with known-item
retrieval, but users often ask questions that are relational or procedural:
\emph{``Where is my food-order chat?''}, \emph{``Which topic holds the notes from
the interview?''}, \emph{``How do I attach a file?''}. These are
questions about the interface and its contents, not about the world.

We use the term \emph{meta access} for an assistant's structured access to the
interface that hosts it: it may access and potentially alter information about
the interface and any of the user's data.
Meta access is related to, but distinct from,
retrieval-augmented generation over documents \cite{lewis2020rag} and tool use
by language models \cite{yao2022react,schick2023toolformer}. The object of
retrieval is the user's own workspace state; the object of action is the
interface itself. Meta access therefore sits at the intersection of
conversational search, command palettes, recommender-style navigation, and
agentic tool use.

\subsection{What ``meta'' means here}

The prefix \emph{meta} comes from Greek and denotes a level that is about,
beyond, or refers to another level. Metadata is data about data; a metamodel is
a model that describes the concepts and relations of another model;
metacommunication is communication about communication. In each case, the
meta-level does not replace the object level. It describes it, constrains it,
or acts on it.

A chatbot normally operates at the object level: it talks about the user's
task, the world, or the content of the current conversation. \emph{Meta access}
adds a meta-level whose object is the chat application itself. The assistant
reasons about conversations, topics, memories, features, and settings rather
than only about the subject of the dialogue. Its context is not the full
interface but a workspace model: a structured description of the entities the
interface exposes (chats, topics, features, preferences), ideally derived from
the application's own metamodel, and, eventually, the operations that can be
requested on them. This is the sense in which the prototype has a
metamodel-like view. It does not reproduce the interface; it uses an abstraction
of the interface's concepts and capabilities.

At the scale of a whole application, this abstraction is an application
metamodel: a description of the application's entities, relations, and
operations. Many applications already have such a model internally---a document
model in an editor, a project and syntax model in an IDE, a schema in a
database, a scene graph in a graphics tool---and debugging or profiling
interfaces already expose parts of it in a controlled way. Meta access can reuse
these models rather than relying on pixels or ad hoc scraping. The design
question is which parts of the metamodel to expose, at what level of stability
and permission, and how to map them into a capability layer that works across
applications.

The distinction separates three things that are easy to confuse. Metadata
access is the narrowest form of meta access: it reads descriptive fields such
as titles, topics, and timestamps. Meta-level knowledge also includes the
feature catalogue, the persona catalogue, and current preference values, which
describe the application rather than a single conversation. Meta-level action
goes further and changes the object level, for example by moving a chat or
changing a memory setting. The ladder in Section~\ref{sec:ladder} follows this
progression from describing the interface, to reading its state, to acting on
it.

We use ``meta'' in this self-referential sense, not in the corporate sense of
Meta Platforms and not as a synonym for the metaverse. It is also different
from meta-learning or meta-prompting: those describe learning or prompting
about learning or prompting, whereas meta access describes an assistant's
access to the application that hosts the interaction.

\subsection{Contributions}

This paper is a position paper grounded in an existing implementation. We make
five contributions:
\begin{enumerate}[leftmargin=1.5em,itemsep=0.2em]
  \item We present a ladder of meta-access levels that separates feature
        tutoring, metadata, settings, and preference access, summary and
        message-level retrieval, confirmed mutating actions, and full
        orchestration, and report how far the implemented prototype reaches on
        that ladder.
  \item We propose meta access as a general design requirement for AI-era
        interfaces, not only chat clients, and show how the ladder, capability
        model, and guardrails transfer to other stateful applications and to
        general-purpose agentic harnesses, whose self-configuration adds a
        separate control-plane problem.
  \item We articulate the opportunities of using a chat interface as a first
        access layer for the whole interface, with search and navigation as the
        first concrete use cases, and define guardrail contracts that tie each
        capability to its evidence, scope, target validation, disclosure,
        confirmation, and failure behaviour.
  \item We outline risks and an evaluation agenda for measuring whether meta
        access actually helps users.
  \item We present the \metachat{} prototype in the Chat AI client as a
        demonstrator of these opportunities and design choices.
\end{enumerate}

\section{Related Work}

Meta access draws on several literatures that rarely meet: conversational
search and personal re-finding; LLM agents, tool use, and interface automation;
retrieval, memory, and context; and mixed-initiative interaction, security, and
human factors. We review them in these groups, compare representative work
along nine dimensions, and then position the present work relative to them.

\subsection{Conversational search and personal re-finding}

Information retrieval has long treated dialogue as an interface to a search
system. Radlinski and Craswell \cite{radlinski2017theoretical} frame
conversational search as a multi-turn collaboration in which system and user
clarify an information need, retrieve results, and interpret them; Trippas et
al.\ \cite{trippas2018informing} derive design implications from spoken
conversational search, and Zamani et al.\ \cite{zamani2022conversational}
survey the field's shift from ad hoc retrieval to interactive problem solving.
This work supplies the interaction vocabulary that meta access inherits:
clarification, follow-up queries, mixed initiative, and results that support
both answer and exploration. Its usual target, however, is an external
collection such as the web or a document corpus, and classical lexical
retrieval, including BM25 and its descendants \cite{robertson2009bm25}, remains
a strong baseline for known-item search.

Personal information management (PIM) studies how people keep, organise, and
re-find their own digital material. Whittaker \cite{whittaker2011personal}
frames PIM as a shift from consumption to curation, and Dumais et al.\
\cite{dumais2003stuff} show that a unified index over email, files, and web
pages, combined with contextual cues such as time and authorship, supports
re-finding; users often remember context rather than exact content, and search
and browsing are complementary. A chat workspace has the same character: it is
small enough to browse but grows large enough to need search, and its items
carry rich contextual metadata. Meta access is a conversational form of
re-finding over that collection. The prototype uses titles, topics, and recency
as metadata cues, summaries as compressed content, and local lexical search
over message text for content questions, bringing it closer to full-text PIM
systems such as ``Stuff I've Seen'' while retaining an interface-aware layer.

Recent surveys extend this vocabulary to LLM-era systems. Mo et al.\
\cite{mo2025conversational} identify query reformulation, clarification,
conversational retrieval, and response generation as core modules, while Zhang
et al.\ \cite{zhang2026agentsurvey} review LLM agents for recommendation and
search. Both reinforce the view that information access is becoming an
interactive, agentic process rather than a single query--response exchange.

\subsection{LLM agents, tool use, and interface automation}

Retrieval-augmented generation grounds model output in retrieved text
\cite{lewis2020rag,gao2023rag}. ReAct \cite{yao2022react}, Toolformer
\cite{schick2023toolformer}, and Gorilla \cite{patil2024gorilla} show that
language models can interleave reasoning with calls to external tools, although
surveys and benchmarks document persistent weaknesses in planning, reliability,
and generalisation \cite{xi2023rise,liu2023agentbench}. A parallel line
automates graphical interfaces: WebGPT \cite{nakano2021webgpt}, Mind2Web
\cite{deng2023mind2web}, WebArena \cite{zhou2023webarena}, OSWorld
\cite{xie2024osworld}, and OSWorld-MCP \cite{jia2025osworldmcp} operate through
browsers, pixels, or GUI actions and are evaluated on open-ended task
completion. Ramineni et al.\ \cite{ramineni2026mcpa11y} expose platform
accessibility trees through a shared MCP schema, so assistive agents need
neither screenshots nor per-platform integrations.

OpenCode is a general-purpose agentic harness rather than an application with a
specific purpose. It runs in the terminal, an IDE, or a desktop application,
supports arbitrary models and tools, and keeps its behaviour in editable
configuration \cite{opencode2026}. Its workspace is whatever task it is given,
usually repositories, shells, or editors, and its control plane includes
prompts, tool and server registries, permission policies, hooks, and subagents.
This generality matters for meta access: there is no single application
metamodel to expose, and if the harness can read or change its own
configuration, the assistant may alter the capability layer that is supposed to
constrain it. We call this reflexive meta access and treat it
separately in Section~\ref{sec:harnesses}. The Model Context Protocol (MCP)
makes this control plane concrete. Hou et al.\ \cite{hou2025mcp} survey its
lifecycle and threat landscape and show that discovery, schemas, and
tool/resource exposure are security decisions made before any model call; for a
harness, the capability registry is therefore an attack surface, not just
configuration.

Meta access is a narrower and more structured form of the same idea. The
environment is the hosting application, the observations are typed metadata
rather than pixels, and the available actions can be declared explicitly
instead of inferred from a GUI. The agent literature is therefore both an
enabler and a caution: it shows that tool use works, but also that long-horizon
autonomy is fragile, which motivates starting with read-only metadata and
preference access and adding confirmed, narrowly scoped actions.

\subsection{Retrieval, memory, and context}

Because a workspace can exceed any context window, meta access depends on how
the interface is compressed. Retrieval-augmented generation
\cite{lewis2020rag,gao2023rag} supplies the pattern of retrieving evidence
before generating an answer. MemGPT \cite{packer2023memgpt} treats an LLM as an
operating system with hierarchical memory; Generative Agents
\cite{park2023generative} maintain a memory stream with reflection; and
LongMemEval \cite{wu2024longmemeval} benchmarks long-term interactive memory.
Long-context models are not a substitute: they can underuse information in the
middle of a long prompt \cite{liu2023lost}, and BEAM
\cite{tavakoli2025beyond} shows that even million-token models struggle as
dialogues lengthen. MemoryAgentBench \cite{hu2025memory} finds that current
memory agents do not master accurate retrieval, test-time learning, long-range
understanding, and selective forgetting; LOCOMO-CONV
\cite{chang2026whenusers} shows that conversational framing exposes retrieval
gaps that QA-style benchmarks miss; and MemUse \cite{sumida2026memuse} reports
a retrieval--integration gap in a four-month deployment. Together these results
argue for a tiered index, query rewriting, implicit-intent handling, and
integration-aware evaluation rather than benchmark accuracy alone.

Conversation summarisation provides one practical compression mechanism.
Surveys of dialogue summarisation \cite{feng2022survey} describe methods for
condensing multi-turn interactions while preserving salient content. The meta
chat uses such summaries together with topics, memories, and per-chat settings;
it adds selective lexical message excerpts when a question needs content. This
keeps cost bounded and reduces exposure compared with shipping full histories,
but summaries are lossy and can be stale, and lexical retrieval depends on the
message wording. Recent memory work makes provenance explicit: Lu et al.\
\cite{lu2026propose} use message-level evidence chains and
propose--verify--commit writes with revision history, a structured counterpart
of confirmed memory mutations and conditional undo. Hierarchical, hybrid, and
semantic retrieval are the natural next steps.

\subsection{Mixed-initiative interaction and human--AI guidelines}

Meta access should not be a one-shot command interpreter. Horvitz
\cite{horvitz1999principles} sets out principles for mixed-initiative
interfaces, including reasoning about uncertainty, the user's goals, and the
cost of asking versus acting; Allen et al.\ \cite{allen1999mixed} discuss the
dialogue-management challenges, and Amershi et al.\ \cite{amershi2019guidelines}
distill widely used guidelines for human--AI interaction. Studies of LLM use
reinforce the need for guidance and verification: non-experts struggle to design
prompts \cite{zamfirescu2023johnny}, users of code-generation tools inspect and
iterate on model output \cite{vaithilingam2022expectation,barke2023grounded},
and natural-language database interfaces documented ambiguity and robustness
problems long before LLMs \cite{androutsopoulos1995natural}. Laban et al.\
\cite{laban2025lost} add a multi-turn caution: tested LLMs perform worse in
multi-turn than single-turn conversations, often commit to assumptions early,
and fail to recover, which makes explicit state and clarification design
requirements. Human-factors studies reinforce this: Schneider et al.\
\cite{schneider2026chatlogs} document linguistic patterns and breakdowns in a
retrieval-augmented assistant, Yu et al.\ \cite{yu2026searchorchat} find no
learning advantage over search when user characteristics are considered, and
Jerry et al.\ \cite{jerry2026accessible} propose a model-based approach to
LLM-driven accessible interfaces. Together they suggest that chat should
complement, not replace, the visual interface, and that meta access must be
judged by user outcomes and breakdowns.

\subsection{Security and privacy of LLM-integrated applications}

Any assistant that reads user content and can act on the user's behalf is a
security boundary. Prompt injection \cite{perez2022ignore} and indirect prompt
injection \cite{greshake2023not} show that untrusted retrieved text can redirect
a model's behaviour; AgentDojo \cite{debenedetti2024agentdojo} benchmarks these
attacks and defences in tool-using agents, and StepJack \cite{zhan2026stepjack}
shows that indirect injection can be decomposed into innocuous steps across a
navigation path. CaMeL \cite{debenedetti2025camel} separates control flow from
data flow and enforces capability checks on data flows, while ClawGuard
\cite{zhao2026clawguard} enforces a user-confirmed rule set at each tool-call
boundary. AC4A \cite{sharma2026ac4a} turns this into access control: instead of
all-or-nothing access to an API or page, an agent receives permissions over
specific resources and actions. AIOS \cite{mei2025aios} places scheduling,
memory, storage, access control, and isolation in an operating-system-like
kernel, moving resource governance from prompts to the platform. PrivacyBench
\cite{mukhopadhyay2026privacybench} benchmarks privacy in personalised AI, and
Wang et al.\ \cite{wang2026hidden} identify irrelevant personalisation,
preference narrowing, and sycophancy as side effects of conditioning on
conversation history.

The implication for meta access is that retrieved content must be treated as
data rather than instructions, tools must be scoped to the user's intent, and
mutations should require confirmation. Privacy is the other side of the
boundary. Meta access sends titles, topics, summaries, and memories to a model
endpoint. This is less exposing than sending full message histories, but it is
not anonymous: summaries and titles can reveal sensitive context. The
local-first design of the host client, in which conversations live in the
browser's indexed database, is a useful starting point, but the boundary between
local state and remote inference must be made explicit to users.

\subsection{Comparative analysis}

Table~\ref{tab:matrix} compares representative related work along nine
dimensions that matter for meta access. The aim is not to rank papers but to
locate the gap that meta access addresses. We use three values: \textbf{+}
means the dimension is explicit and central to the paper; \textbf{0} means it
is partial, indirect, or limited to a narrow setting; \textbf{-} means it is
not addressed or is out of scope. The ratings describe each paper's stated
scope, not the quality of its contribution.

\begin{table}[htbp]
\centering
\footnotesize
\setlength{\tabcolsep}{2.5pt}
\renewcommand{\arraystretch}{1.02}
\begin{tabular}{@{}l|*{3}{c}|*{3}{c}|*{3}{c}|}
\toprule
\textbf{Work}
& \rotatebox[origin=c]{90}{\textbf{Dialogue}} 
& \rotatebox[origin=c]{90}{\textbf{Personal}}
& \rotatebox[origin=c]{90}{\textbf{Metadata}}
& \rotatebox[origin=c]{90}{\textbf{Content}}
& \rotatebox[origin=c]{90}{\textbf{Interface}}
& \rotatebox[origin=c]{90}{\textbf{Actions}}
& \rotatebox[origin=c]{90}{\textbf{Grounding}}
& \rotatebox[origin=c]{90}{\textbf{Security}}
& \rotatebox[origin=c]{90}{\textbf{User study}} \\
\midrule
\multicolumn{10}{@{}l@{}}{\emph{Conversational search}} \\
Radlinski \& Craswell \cite{radlinski2017theoretical}
  & + & - & - & + & - & - & 0 & - & - \\
Trippas et al.\ \cite{trippas2018informing}
  & + & - & - & + & - & - & 0 & - & + \\
Zamani et al.\ \cite{zamani2022conversational}
  & + & - & 0 & + & - & - & 0 & 0 & - \\
Mo et al.\ \cite{mo2025conversational}
  & + & 0 & 0 & + & - & - & 0 & 0 & - \\
Zhang et al.\ \cite{zhang2026agentsurvey}
  & + & 0 & 0 & + & 0 & + & 0 & 0 & - \\
\midrule
\multicolumn{10}{@{}l@{}}{\emph{Personal information management}} \\
Dumais et al.\ \cite{dumais2003stuff}
  & - & + & + & + & - & - & 0 & 0 & + \\
Whittaker \cite{whittaker2011personal}
  & - & + & 0 & + & - & - & - & 0 & - \\
\midrule
\multicolumn{10}{@{}l@{}}{\emph{Retrieval and agents}} \\
Lewis et al.\ \cite{lewis2020rag}
  & - & - & - & + & - & - & + & - & 0 \\
Yao et al.\ \cite{yao2022react}
  & - & - & - & 0 & 0 & + & 0 & - & - \\
Schick et al.\ \cite{schick2023toolformer}
  & - & - & - & 0 & 0 & + & 0 & - & - \\
Patil et al.\ \cite{patil2024gorilla}
  & - & - & + & 0 & 0 & + & 0 & - & - \\
Deng et al.\ \cite{deng2023mind2web}
  & - & - & - & 0 & + & + & + & 0 & 0 \\
Zhou et al.\ \cite{zhou2023webarena}
  & - & - & - & 0 & + & + & 0 & - & 0 \\
Xie et al.\ \cite{xie2024osworld}
  & - & - & - & 0 & + & + & 0 & - & 0 \\
Jia et al.\ \cite{jia2025osworldmcp}
  & - & - & - & 0 & + & + & 0 & 0 & - \\
Mei et al.\ \cite{mei2025aios}
  & - & - & 0 & 0 & - & + & - & 0 & - \\
Ramineni et al.\ \cite{ramineni2026mcpa11y}
  & - & 0 & + & 0 & + & + & 0 & 0 & - \\
\midrule
\multicolumn{10}{@{}l@{}}{\emph{Memory and long context}} \\
Packer et al.\ \cite{packer2023memgpt}
  & + & + & 0 & + & - & 0 & 0 & - & - \\
Park et al.\ \cite{park2023generative}
  & + & + & + & + & - & + & 0 & - & + \\
Tavakoli et al.\ \cite{tavakoli2025beyond}
  & + & 0 & 0 & + & - & - & 0 & - & 0 \\
Hu et al.\ \cite{hu2025memory}
  & + & 0 & 0 & + & - & 0 & 0 & - & - \\
Chang \& Chen \cite{chang2026whenusers}
  & + & 0 & 0 & + & - & - & + & - & 0 \\
Sumida et al.\ \cite{sumida2026memuse}
  & + & + & 0 & + & - & - & 0 & 0 & + \\
\midrule
\multicolumn{10}{@{}l@{}}{\emph{Interaction and security}} \\
Horvitz \cite{horvitz1999principles}
  & 0 & 0 & - & - & + & + & - & - & 0 \\
Amershi et al.\ \cite{amershi2019guidelines}
  & 0 & 0 & - & - & + & 0 & 0 & 0 & + \\
Laban et al.\ \cite{laban2025lost}
  & + & - & - & 0 & - & - & - & - & - \\
Greshake et al.\ \cite{greshake2023not}
  & 0 & 0 & - & + & 0 & + & - & + & - \\
Debenedetti et al.\ \cite{debenedetti2024agentdojo}
  & 0 & 0 & - & + & 0 & + & - & + & - \\
Sharma \& Grossman \cite{sharma2026ac4a}
  & - & 0 & - & + & + & + & - & + & - \\
Zhan et al.\ \cite{zhan2026stepjack}
  & 0 & 0 & - & + & + & + & - & + & - \\
Schneider et al.\ \cite{schneider2026chatlogs}
  & + & 0 & 0 & + & 0 & - & 0 & 0 & + \\
Yu et al.\ \cite{yu2026searchorchat}
  & + & - & - & + & 0 & - & 0 & - & + \\
\midrule
\multicolumn{10}{@{}l@{}}{\emph{This work}} \\
\metachat{} prototype
  & + & + & + & + & + & + & + & + & - \\
\bottomrule
\end{tabular}
\caption{Comparative matrix of representative related work. \textbf{+}:
explicitly addressed and central; \textbf{0}: partial, indirect, or limited;
\textbf{-}: not addressed or out of scope. Ratings describe scope, not
quality. This work implements selective message-level content retrieval and
confirmed, reversible actions; the user-study column remains out of scope.}
\label{tab:matrix}
\end{table}

Three patterns stand out. First, conversational search and PIM occupy opposite
corners: conversational search supports dialogue and content retrieval but not
the user's personal workspace, while PIM supports personal metadata and content
but not dialogue or interface actions. Second, agent and GUI-automation work
supports actions and interface operation, but its object is an open environment
rather than the user's own conversation history; personal scope, structured
workspace metadata, and user-facing grounding are typically absent. Third,
memory and long-context work supports dialogue and personal context, but it
rarely models the interface or provides a capability layer for acting on it.

A trade-off is also visible: many systems that act do not expose citations to
the underlying workspace items, while systems with strong grounding often do
not act. Security work supplies adversarial benchmarks and defences but is
rarely integrated with personal retrieval or interface help, and HCI work
supplies interaction principles but not a concrete architecture. Meta access is
the region that remains sparsely populated: personal scope, structured
metadata, interface awareness, and grounding. The prototype now adds selective
message-level retrieval and confirmed, reversible mutations, while the ladder in
Section~\ref{sec:ladder} makes the remaining boundaries explicit: hierarchical
retrieval, broader level-4 actions, durable audit, and level-5 orchestration.

Existing work covers conversational search over external corpora, personal
re-finding over files and email, and tool-using agents on web pages or operating
systems. Meta access sits between these lines: the object of retrieval is the
application's own state, and the environment is a declared capability layer
rather than an open GUI. The contribution of this paper is to name that
combination, demonstrate it with confirmed, reversible actions, propose a ladder
from tutoring to orchestration, and set out an evaluation agenda.

\section{The Meta Chat Prototype}

\subsection{Placement and interaction model}

The \metachat{} is a separate chat in Chat AI \cite{doosthosseini2026saia}.
It is opened from an AI icon in the sidebar header and reuses the normal chat
shell, so it feels like the rest of the application rather than a bolted-on
assistant. A banner above the conversation states its two purposes---learning
how to use Chat AI and finding past chats and topics---and offers entry points
to browse tutorials and use cases, which helps users who do not yet know what
to ask. In short, the meta chat is a normal conversation with special routing
and special privileges, not a separate service.

\subsection{What the assistant sees}

The meta assistant can access every conversation and read every user-facing
setting or control, so it can answer questions about the workspace as a whole
rather than about a pre-selected subset. At the metadata level, the client
supplies a structured, size-bounded view: conversation identifiers, titles,
topics, recency, and compact configuration descriptions; the topic list, current
setting and preference values, saved memories, the model catalogue, and the
\metachat{}'s own settings. A bounded context is therefore a packaging choice,
not a restriction on what the assistant may access.

At the content level, the client adds short conversation summaries and, when a
question needs exact evidence, searches all stored message text locally and adds
only the most relevant excerpts, each tied to a citable message. Decision and
comparison questions also receive a few recent excerpts. Summaries are generated
out of band when a conversation is left, so they can be missing, incomplete, or
stale; the prompt tells the model not to invent them.

This is a deliberate trade-off. It keeps context cost bounded and exposes less
message content than a full-history baseline, but recall is limited: details
absent from both summaries and selected excerpts remain invisible, and lexical
search depends on wording. The prototype is therefore a hybrid of metadata
access, summary and selective message retrieval, and confirmed mutation rather
than a hierarchical or semantic retriever.

\subsection{Grounding references and feature help}

The system prompt constrains the assistant in two ways. References must use
internal links whose identifiers come only from the supplied context: links to a
conversation, an exact message, a topic, a persona, or a settings tab. The
renderer recognises these routes and navigates inside the application, scrolling
to and briefly highlighting the target. References therefore act as direct
routes into the workspace, not mere citations. The assistant must also stay
within the supplied context and must not invent conversations, topics, messages,
memories, personas, or preference values.

The assistant additionally receives a machine-readable catalogue of interface
features, each with a description, keywords, and guided-tour steps that target
real controls. It can explain a feature in prose and attach a marker that the
client turns into an optional step-by-step tour. Feature help is therefore not
only textual: it can hand the user over to the relevant part of the interface.

\subsection{Tutorials, settings, preferences, and navigation}

\begin{figure}[htbp]
  \centering
  \includegraphics[width=\linewidth]{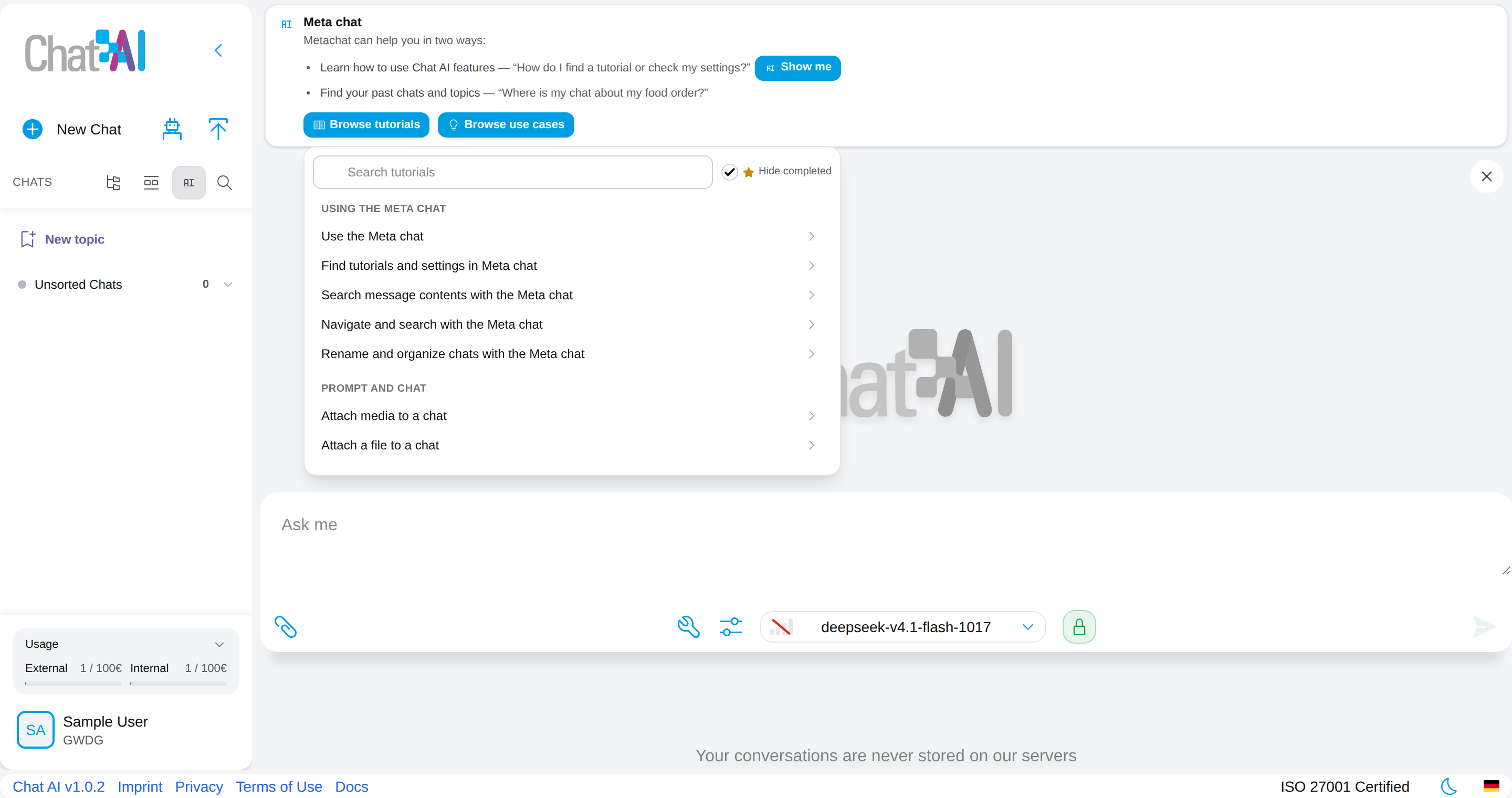}
  \caption{The tutorial library, opened from the \emph{Browse tutorials} entry
  point in the \metachat{} banner. Tutorials are grouped by task and can be
  started directly; each runs a guided tour of the real controls.}
  \label{fig:tutorials}
\end{figure}

\begin{figure}[htbp]
  \centering
  \includegraphics[width=\linewidth]{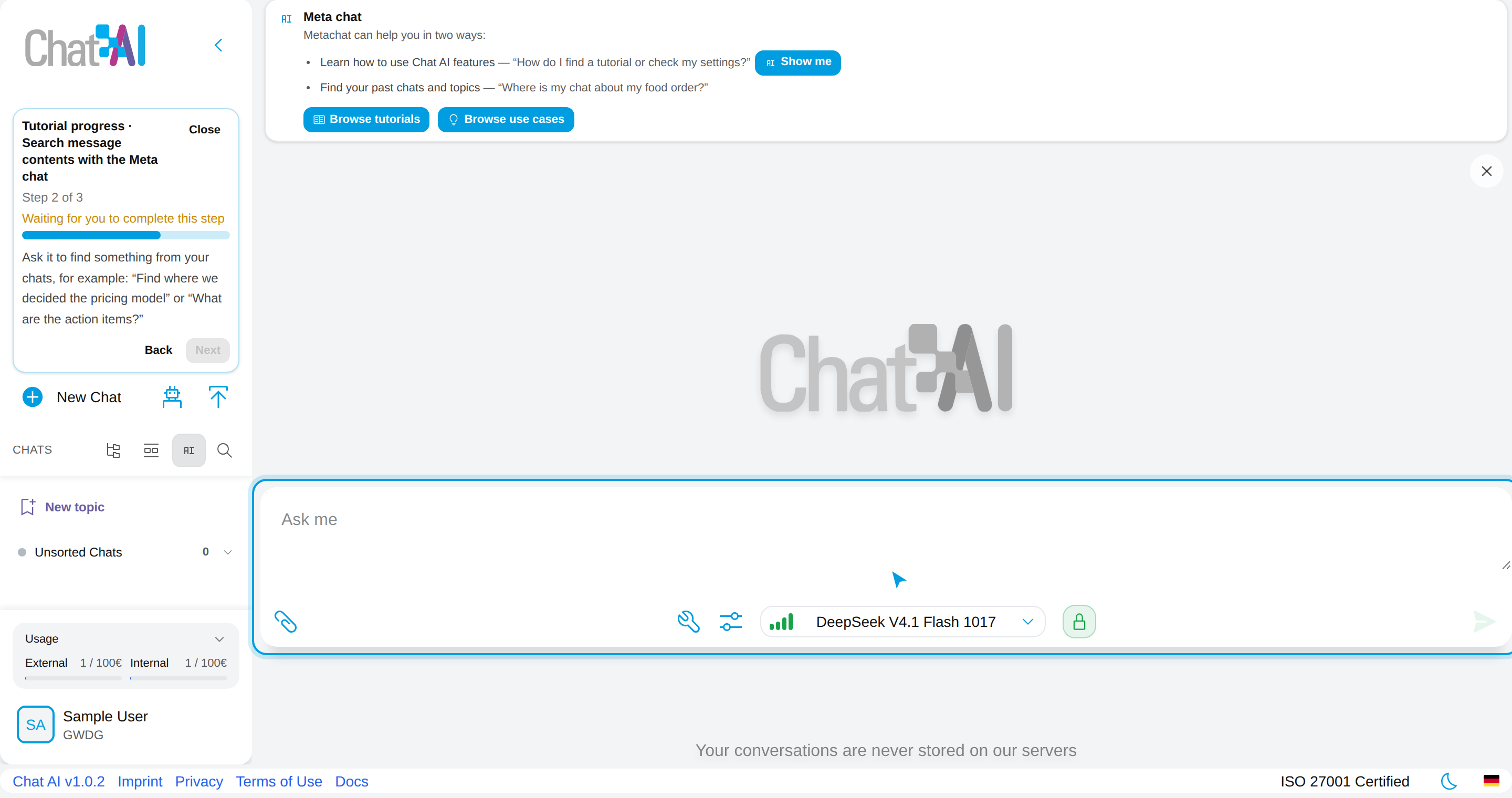}
  \caption{Running a tutorial in the \metachat{}, captured from the local
  application. The tutorial status panel shows step 2 of 3, names the expected
  action, and highlights the prompt area. Each step waits for the corresponding
  interface action or message completion before advancing.}
  \label{fig:tutorial-running}
\end{figure}

The feature catalogue also powers a browsable tutorial library, reachable from
the banner or by name. This turns feature help from something the user must know
how to ask for into something they can browse, which matters for unfamiliar or
infrequently used capabilities. Figures~\ref{fig:tutorials} and
\ref{fig:tutorial-running} show the library and a tutorial in progress.

\begin{figure}[htbp]
  \centering
  \includegraphics[width=\linewidth]{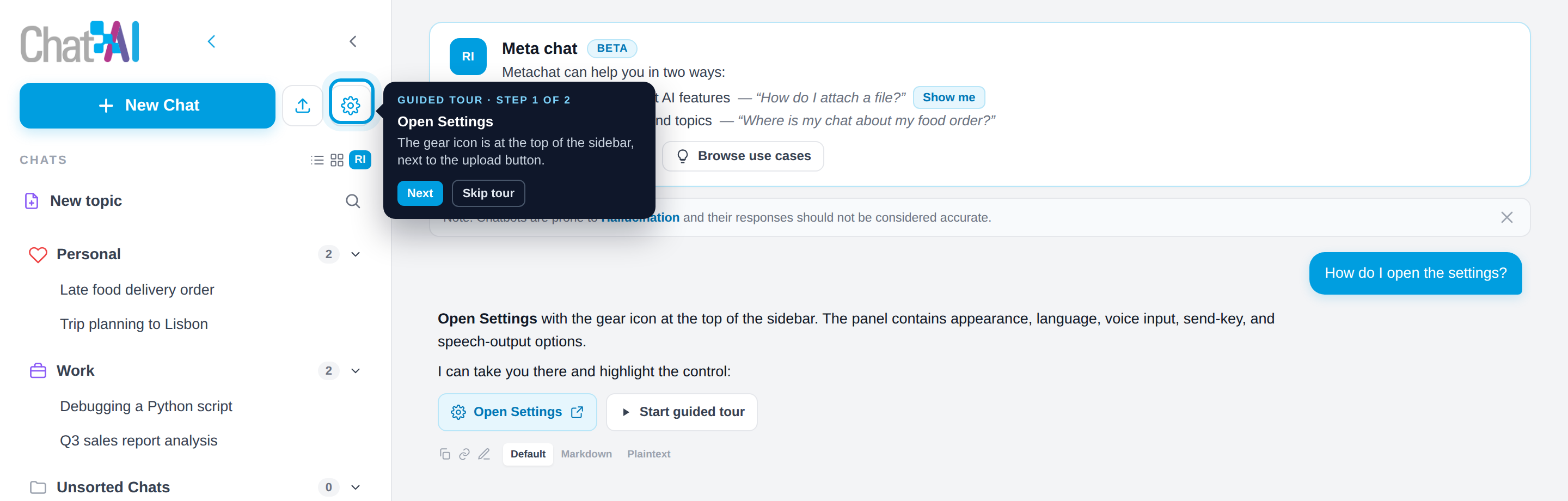}
  \caption{Routing a settings question to the interface, shown with synthetic
  data. The answer names the control, offers to open the settings panel, and
  starts a tour that highlights the gear icon. Navigation remains read-only;
  changing the value is a separate confirmed proposal.}
  \label{fig:settings}
\end{figure}

Guidance is not limited to prose. For a settings question, the assistant can
explain where the control is and offer to open the relevant panel and highlight
it (Figure~\ref{fig:settings}). Opening a panel is navigation; changing a value
is a separate mutation that the user must confirm.

\begin{figure}[htbp]
  \centering
  \includegraphics[width=\linewidth]{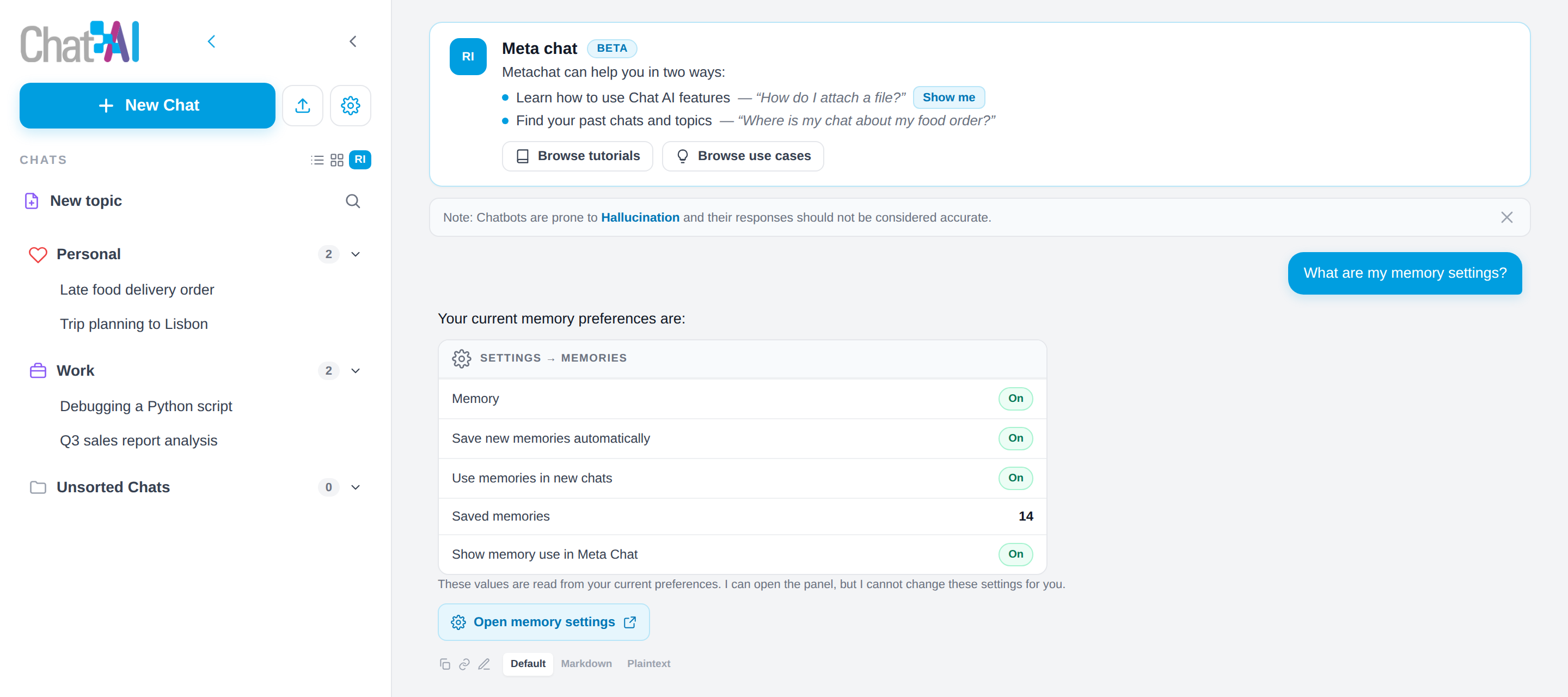}
  \caption{Reading current preference values in the \metachat{}, shown with
  synthetic data. The answer prints the user's memory settings, offers to open
  the corresponding settings page, and can propose a value change. A change is
  applied only after an explicit confirmation and can be reverted while the
  resulting state still matches.}
  \label{fig:preferences}
\end{figure}

Meta access also covers the current value of any preference, not only the
location of its control. For a memory question, the assistant reports the
current configuration and offers to open the corresponding settings page
(Figure~\ref{fig:preferences}). A proposed change is shown as a warning card and
is applied only after explicit confirmation.

\begin{figure}[htbp]
  \centering
  \includegraphics[width=\linewidth]{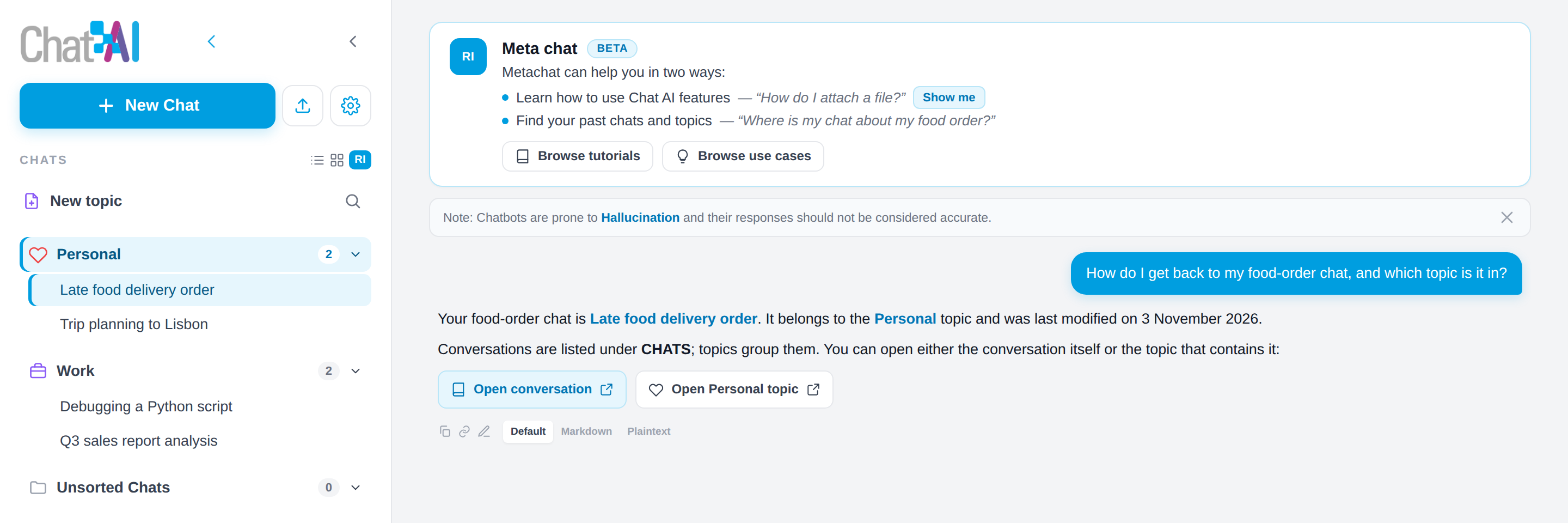}
  \caption{Navigating conversations and topics in the \metachat{}, shown with
  synthetic data. The answer links to the cited conversation and to its topic;
  the sidebar highlights both targets so the user can see where the link will
  lead.}
  \label{fig:navigation}
\end{figure}

The same link vocabulary covers the workspace itself: a reference can point to a
conversation, an exact message, or the topic that contains it, so the user can
open either the item or its surrounding context. This is particularly useful for
re-finding, where a title alone may be ambiguous but its topic supplies
orientation and a message citation resolves the exact statement
(Figure~\ref{fig:navigation}).

\FloatBarrier

\subsection{Persona catalogue and handoff}

Personas extend meta access beyond the user's conversations to a curated
configuration space. A persona bundles a title, description, model, sampling
parameters, and a system prompt. The sidebar's \emph{Chat with Persona} action
opens a library grouped by category; selecting an entry imports it as a new
chat and applies its settings.

\begin{figure}[htbp]
  \centering
  \includegraphics[width=\linewidth]{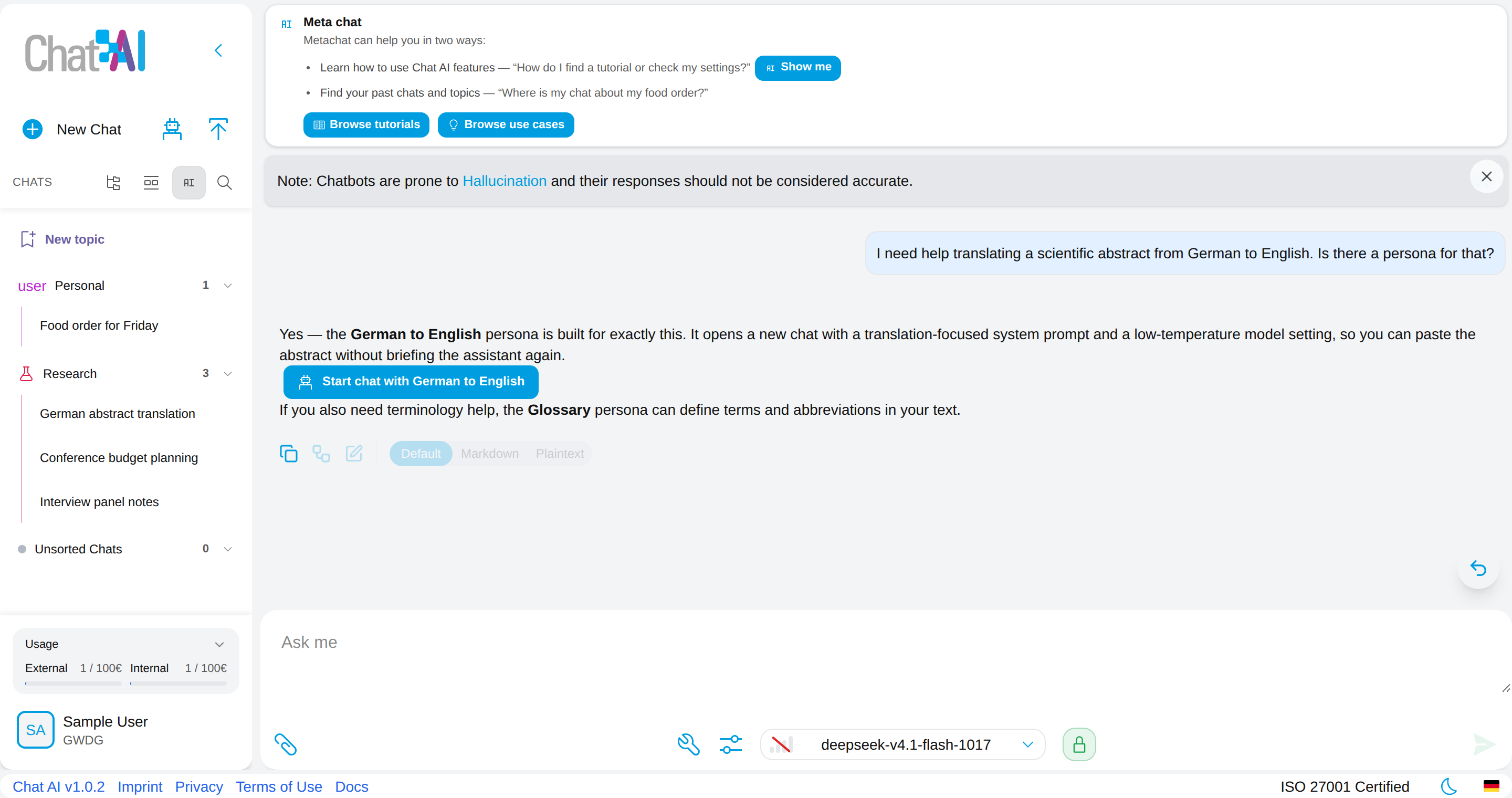}
  \caption{Persona recommendation and handoff in the \metachat{}, shown with
  synthetic data. The assistant recommends a persona from catalogue metadata
  and offers a start link; clicking it opens a new chat with the persona's
  title, model, and system prompt. The \metachat{} itself never reads the
  persona prompt.}
  \label{fig:personas}
\end{figure}

The \metachat{} treats the catalogue as metadata. When a request matches a
persona, the assistant names it, explains why it fits, and emits a link that the
renderer turns into a start action. Clicking is an explicit user decision: the
client fetches the definition, creates the conversation, and applies the
configuration (Figure~\ref{fig:personas}). The assistant neither reads the
prompt nor executes the import, so the handoff does not give it control over the
definition.

This division is deliberate. It gives the user a one-click path from an intent
to a configured conversation without exposing an untrusted system prompt to the
assistant or letting the model change settings on its own. The cost is that the
recommendation is only as good as the catalogue's metadata: a poorly described
persona is hard to find, and a stale entry can promise behaviour it no longer
implements.

\subsection{Persistence, privacy, and scope}

Because the meta conversation is stored in the same local database as ordinary
chats, meta access inherits the client's local-first model: conversation data
stays in the browser, and the hidden meta conversation can be exported and
re-imported. At the same time, the meta flow sends titles, topics, summaries,
selected excerpts, memories, persona metadata, the model catalogue, and relevant
setting and preference values to the configured model endpoint. Metadata is less
sensitive than complete messages, but selected excerpts are still message
content, and titles, summaries, memories, and settings can reveal a great deal.
Starting a persona also fetches its definition from an external repository, a
request that should be visible in the interface.

The assistant can read and propose changes to essentially any control or
setting, including conversation organisation, preferences, per-chat
configuration, and memories. It never executes a mutation directly. Every
mutating change is presented in the UI as a proposal, and the client applies it
only after the user explicitly consents. After applying, revert is allowed only
while the current state still matches what was applied. A visible audit log
remains future work.

\section{A Ladder of Meta Access}
\label{sec:ladder}

Table~\ref{tab:levels} separates capabilities that are often conflated in
discussions of ``chat with your app''. Each level adds power and risk. The
ladder is not specific to chat: it applies to any stateful application that
exposes a workspace model and a capability layer. It assumes, however, an
application with a bounded purpose and a host that enforces the capability layer
from outside the assistant; Section~\ref{sec:harnesses} relaxes both assumptions
for general-purpose agentic harnesses. The key
boundary is mutation: levels 1--3 read, explain, and navigate, whereas level 4
is the first level that changes workspace state through the assistant,
including settings and preferences. Level 5 composes such changes. Within the
read-only levels, level 2 stays at metadata about conversations, whereas level 3
crosses into conversation content, first through summaries and then through
exact message spans. A user-invoked handoff, such as clicking a recommended
persona, is a boundary case: it creates a new chat, but the assistant only
retrieves and links, and the user performs the import. The guardrail contracts
in Section~\ref{sec:guardrails} apply at every level: evidence and target
validation for levels 1--2, provenance and scope for level 3, and confirmation,
reversibility, and audit for levels 4--5. The implemented prototype now covers
levels 1--3 and a substantial part of level 4: it searches message text with
citations and performs confirmed, reversible changes to chats, settings, and
memories. Level 5 remains future work, although the prototype can already group
several proposed changes into one confirmation card and apply a setting to all
chats.

\begin{table}[htbp]
\centering
\small
\renewcommand{\arraystretch}{1.25}
\setlength{\tabcolsep}{4pt}
\begin{tabular}{@{}>{\raggedright\arraybackslash}p{0.05\linewidth}>{\raggedright\arraybackslash}p{0.28\linewidth}>{\raggedright\arraybackslash}p{0.31\linewidth}>{\raggedright\arraybackslash}p{0.24\linewidth}@{}}
\toprule
\textbf{L} & \textbf{Capability} & \textbf{Example} & \textbf{Risk} \\
\midrule
0 & No meta access: only the current dialogue & Ordinary chat interfaces & None new \\
1 & Feature tutoring: describe the interface and run tutorials & ``How do I attach a file?'' & Wrong instructions \\
2 & Metadata access across all conversations: titles, topics, settings, preferences, controls, and personas; navigation & ``Where is my food-order chat?'' & Wrong target; preference disclosure; misleading catalogue entry \\
3 & Conversation content: summaries and message-level retrieval & ``What did we decide about the budget?'' & Sensitive content exposure; stale summaries \\
4 & Confirmed mutating actions: edit chats, topics, or settings & ``Turn off automatic memory saving.'' & Unintended edits or setting changes \\
5 & Multi-step, autonomous orchestration & ``Tidy my topics and rename weak titles.'' & Compounding errors, loss of control \\
\bottomrule
\end{tabular}
\caption{A ladder of meta access. The current \metachat{} implements levels
1--3 and confirmed level-4 mutations for chats, settings, and memories, with
limited composition across several proposals; full level-5 orchestration
remains future work. The levels describe autonomy over workspace state; for
self-configuration in general-purpose harnesses they are combined with a target
class (Section~\ref{sec:harnesses}).}
\label{tab:levels}
\end{table}

\subsection{Level 1: tutoring}

Feature tutoring is the lowest-risk form of meta access and the one most
directly implemented. The assistant explains where a control is, what it does,
and in what order to use it. The guided-tour mechanism closes the loop between
explanation and interface by highlighting the relevant element. The tutorial
library extends the same mechanism from on-demand help to browsable
instruction: users can see which tours exist and start one without first
formulating a question. This addresses a persistent problem in feature-rich
applications: users do not know what they do not know. A chat that can answer
``what can I do here?'' lowers the cost of discovery without requiring the
user to leave the conversation. Figure~\ref{fig:tutorials} shows the
browsable tutorial entry point that implements this level.

\subsection{Level 2: metadata, settings, and preference access}

Level 2 is read-only and stays at metadata about conversations and the
interface; it does not read conversation content. Metadata retrieval supports
questions about the whole workspace: which
conversations exist, their titles and topics, when they were last touched, and
where to find them. This is the prototype's ``where is my food-order chat?''
use case. The answer is useful only if it is grounded
and actionable: the assistant must cite a real conversation or topic and provide
a link that navigates to it. Grounding turns a plausible-sounding answer into a
verifiable one, and the navigation link turns retrieval into a one-click
outcome. The prototype also links to settings and reads the current values
behind any control, so a preference or configuration question can be answered in
the conversation without opening the panel. These are still read-only moves:
the assistant opens a view or reports a value, but the user makes the change.
Figure~\ref{fig:navigation} shows a metadata answer that links both the
conversation and its surrounding topic.

Persona recommendation sits at the same level but extends retrieval to a
catalogue the user did not create. The assistant sees only titles, categories,
and descriptions, and recommends an entry when one clearly matches the request.
The link it emits is a user-invoked handoff: clicking it imports the full
definition and starts a new chat. The assistant does not read the persona
prompt, choose the model, or apply the settings itself. This keeps the
recommendation separate from the user's explicit import decision while turning
it into a concrete starting point. It also means that persona quality is a
retrieval problem: the catalogue must be well described, fresh, and free of
duplicate or misleading entries.

Metadata retrieval is also a natural fit for a conversational command palette.
A user can ask in their own words, in their own language, without knowing the
exact title. The assistant can reason over titles, topic names, and recency, and
can explain why it chose a result. This is more forgiving than substring search,
but also less predictable, which is why citations and links matter.

\subsection{Level 3: conversation content and message-level retrieval}

\begin{figure}[htbp]
  \centering
  \includegraphics[width=\linewidth]{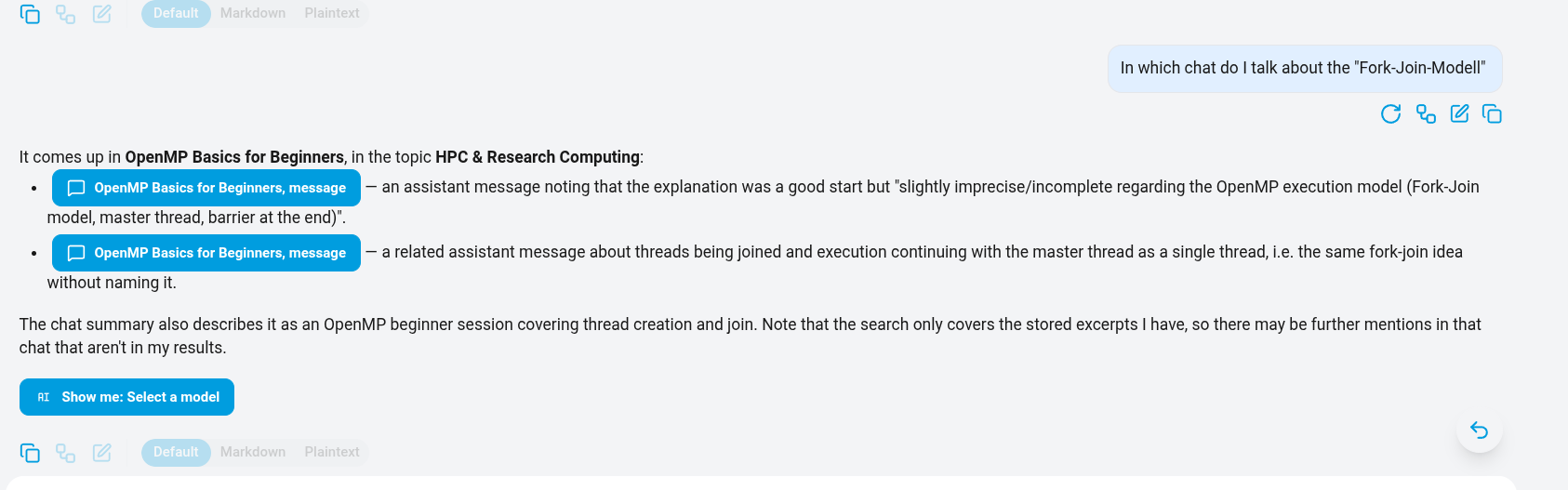}
  \caption{Message-level retrieval in the \metachat{}, shown with synthetic
  data. The user asks where the ``Fork-Join-Modell'' was discussed; the answer
  identifies the chat \emph{OpenMP Basics for Beginners} and offers exact-message
  citations that open and highlight the source.}
  \label{fig:fulltext-search}
\end{figure}

Metadata is a lossy index. Some content questions, such as \emph{``What did we
decide about the budget?''}, may be answered from a summary; others, such as
\emph{``Where was the Fork-Join-Modell discussed?''}, need the exact message.
Summaries provide a compressed first form of content access, but they may omit
the exact statement. The prototype therefore implements message-level retrieval
as well: it runs a local lexical search over stored message text using the
current user message, adds the most relevant excerpts with conversation and
message identifiers to the context, and renders citations that open and
highlight the exact message. For decision, action-item, and comparison
questions it also adds recent excerpts from the most recent conversations.
This extends the prototype beyond summary-only retrieval while keeping context
bounded and exposing only selected spans. Figure~\ref{fig:fulltext-search}
shows the second kind of query resolving to a chat and two exact-message
citations.

The current implementation is deliberately simpler than the principled design
we propose for the next step. It does not yet narrow candidates through a
metadata hierarchy, use embeddings or hybrid semantic search, or relax recency
truncation for older chats. Those changes would improve recall and make the
retrieval depth depend on the question rather than on a single global cap. The
implemented path already provides the citation discipline and the clear
message-level boundary that such a retriever needs.

\subsection{Level 4: confirmed mutating actions}

Level 4 is the first level that changes workspace state. Unlike the read-only
links and preference printouts of level 2, a confirmed action performs a
mutation. The prototype implements a first version: it renames or moves a
conversation, creates a topic as part of a move, changes global preferences,
changes per-chat settings for one or all chats, and adds, edits, deletes, or
clears memories. Per-chat settings include the model, temperature, top\_p,
reasoning effort, memory-mode override, tools, and web-search flags. Settings
count as mutations, as do memory changes.
Opening the settings panel remains navigation (level 2); changing a value is a
mutation (level 4).

\begin{figure}[htbp]
  \centering
  \includegraphics[width=\linewidth]{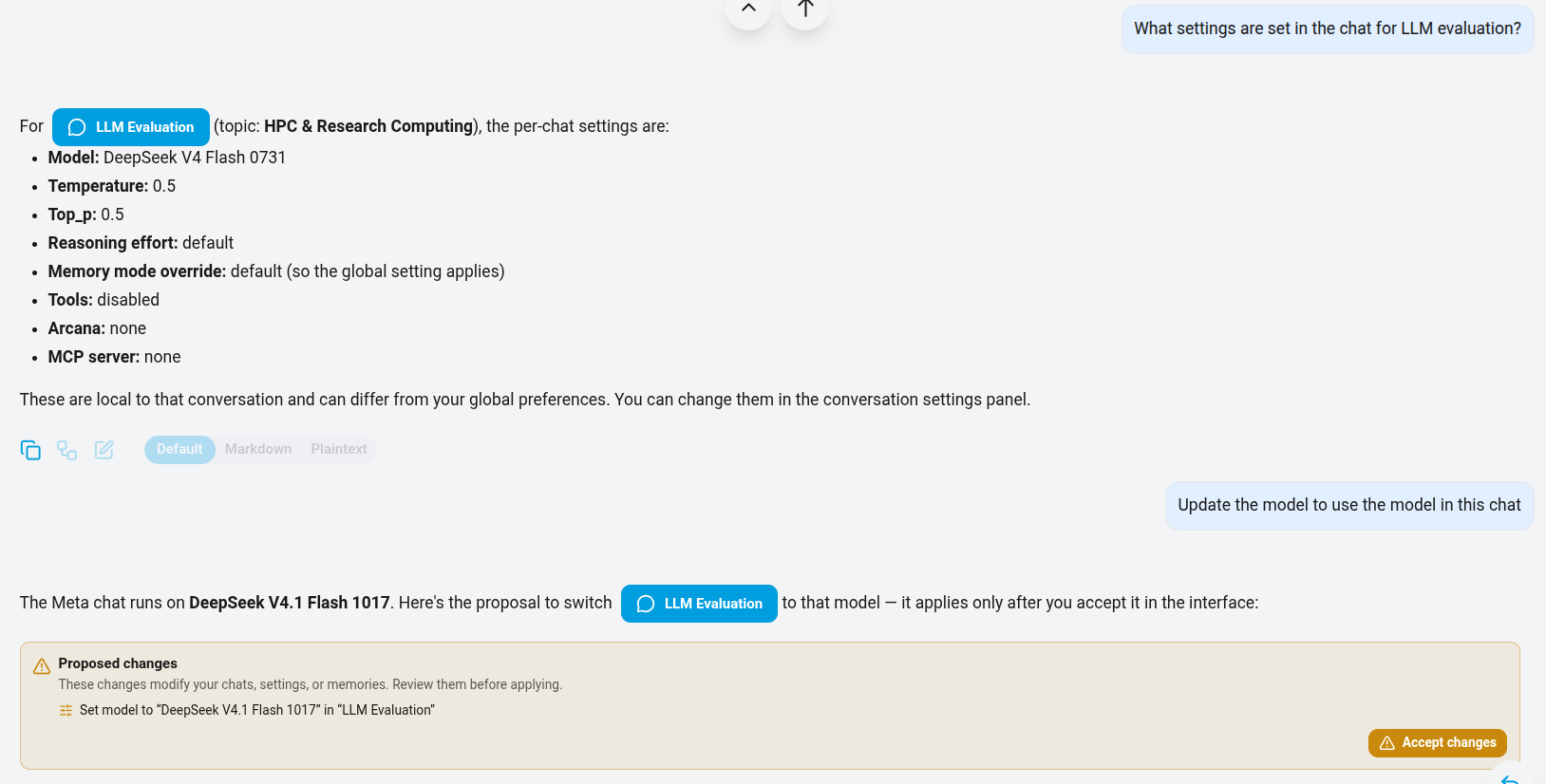}
  \caption{Reading and changing per-chat settings in the \metachat{}, shown with
  synthetic data. The assistant reports the model and temperature of the
  \emph{LLM Evaluation} chat, then proposes a model change in a warning card
  with an explicit \emph{Accept changes} action.}
  \label{fig:changing-settings}
\end{figure}

\begin{figure}[htbp]
  \centering
  \includegraphics[width=\linewidth]{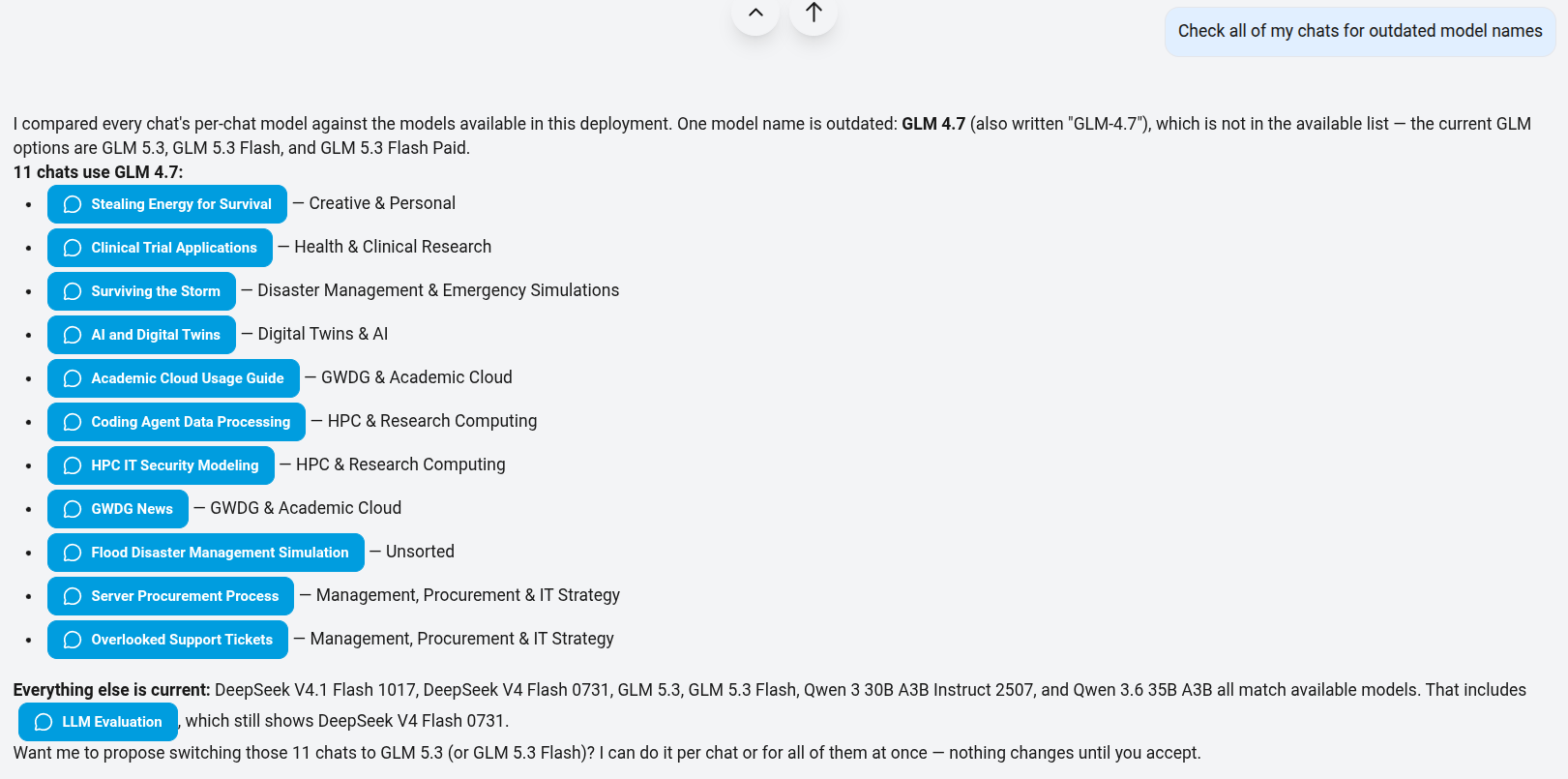}
  \caption{Auditing model use across the workspace in the \metachat{}, shown with
  synthetic data. The assistant identifies chats still using an outdated model
  and presents their titles as links, supporting a later bulk update after the
  user reviews the affected conversations.}
  \label{fig:outdated-model}
\end{figure}

These actions are expressed as typed, narrowly scoped proposals and validated
on the client. The assistant never executes them directly. Instead, the card
below the message shows the target object, the old and new values where
available, a warning icon, and an explicit accept action. After applying, a
revert action is available only if the current state still matches what was
applied; the check prevents overwriting intervening manual edits. This provides
confirmation and conditional reversibility. Still missing are a durable audit
log, topic rename and deletion, chat deletion and export, and archival
operations. Figure~\ref{fig:changing-settings} shows the read, propose, and
confirm cycle for per-chat settings, and Figure~\ref{fig:outdated-model} shows a
workspace-wide model audit that precedes a bulk change.

Confirmation is not a bureaucratic extra. It protects against the failure mode
that makes agentic interfaces frustrating: a confident but wrong action on
personal data. A good interaction shows the proposed change, the target object,
and the expected effect before execution. For bulk operations, a preview is
essential; the prototype lists every proposed change and supports applying the
same per-chat setting to all chats.

\subsection{Level 5: orchestration}

At the top of the ladder, the assistant composes multiple actions: ``group the
chats from last week by project, rename generic titles, and archive the
duplicates.'' This is the most powerful and the most dangerous level. It
requires planning, intermediate verification, the ability to stop and ask, and
a strong notion of scope. The prototype already lets the model place several
typed changes---chat moves, renames, settings, and memories---into one
confirmation card, and it can apply a per-chat setting to all chats. That is
only composition with a single confirmation step, not full orchestration. We
expect level 5 to be appropriate only with permissioning, dry runs, and the
ability to review a diff of proposed changes.

\subsection{Reflexive meta access in agentic harnesses}
\label{sec:harnesses}

The ladder assumes an application with a bounded purpose and a host that
enforces the capability layer from outside the assistant. A general-purpose
agentic harness such as OpenCode \cite{opencode2026} violates both assumptions.
It is not a chat client with a fixed workspace: it is a runtime for arbitrary
tasks, repositories, models, and tools. Its state is the task environment, and
its own configuration---model and prompt selection, tool and server registries,
permission and sandbox policy, hooks, subagents, and project instructions---is a
control plane rather than a purpose-specific application metamodel.

Mechanically, self-configuration still traverses the ladder: reading
configuration, logs, or session metadata is level 2 or 3; a single setting
change is level 4; and composing persistent changes across model, tools, and
permissions is level 5. The risk is not captured by the autonomy level alone.
We therefore distinguish the target of a change from the level of autonomy. A
behavioural setting (verbosity or temperature) is ordinary level 4; a state
setting (memory or retention) carries level-3 privacy consequences; a capability
setting (tools, servers, plugins, or network access) changes what the assistant
can do in future; and an authority setting (permissions, approval policy,
sandbox, or hooks) changes who can approve actions and what can be observed. The
last two are control-plane mutations, and the last is self-escalation.

We call confirmed self-configuration \emph{level 4S} (self) and persistent,
autonomous reconfiguration \emph{level 5S}. These labels combine a level
with a target; they are not new rungs on the same axis. Level 4S with a
behavioural target is mild, whereas level 4S with an authority target is more
dangerous than many level-5 workspace actions. A harness must not let the
assistant approve its own authority changes at any level. Runtime enforcement
frameworks such as ClawGuard \cite{zhao2026clawguard} locate the check at the
tool-call boundary, which is where a harness must enforce capability and
authority scopes. The assistant may propose a diff, but a separate policy layer
or a human should apply it out of band after showing the effect on current and
future sessions. Authority,
capability, and persistent settings need an immutable audit trail, scoped
activation, a dry run, and clear rollback semantics; changes that take effect
only after a restart must say so.

A further complication is that harness configuration often lives in ordinary
files. If the harness also exposes generic file-editing or shell tools, the
assistant can change its settings without using a typed capability at all,
bypassing the confirmation, validation, and audit that the ladder assumes. The
classification therefore has to follow the capability actually used, not the
file path or the setting's name. The safe design is to keep the authority and
capability control plane outside the agent's writable scope, or to treat edits
to it as a distinct, strongly guarded capability.

\section{Why Meta Access Matters Across Interfaces}

\subsection{Search versus find}

Users often do not have a query; they have an intention. Classical search
assumes the user can express a keyword. Conversational meta access lets the
user describe the memory instead: \emph{``the chat where we compared two travel
options''}. The assistant can map that description onto titles, summaries, and
topics, ask a clarifying question if needed, and return a small set of
candidates with links. This is retrieval as a dialogue rather than a one-shot
query, and it can recover from ambiguity instead of returning an empty result
page.

\subsection{Navigation and orientation}

A chat client with hundreds of conversations is a spatial interface: users
remember where things were. Meta access can answer orientation questions
(``what did I work on last week?'', ``which topics are unused?'') and navigate
directly to a target. It can also explain the organisation of the workspace,
which is otherwise implicit in a sidebar tree. This is particularly valuable
for infrequent users, who must re-learn the layout each time they return.

\subsection{Feature discovery and onboarding}

Feature-rich applications accumulate controls that most users never touch.
A meta assistant that knows the feature catalogue can act as an always-available
onboarding layer: it explains, demonstrates, and guides. Because the catalogue
is data rather than hard-coded prose, it can evolve with the interface, and
because the guidance is anchored to real elements, it can be verified by
running the tour. The prototype's tutorial library and feature guides are a
concrete instance of this idea: the user can browse what is available or ask
for help in natural language. The persona catalogue extends discovery from
features to configurations: a user who describes a goal can be handed a
ready-made setup instead of being told which controls to change.

\subsection{Composition across features}

The most interesting potential is composition. A single request can span
retrieval, navigation, and configuration: \emph{``Find the chat about the
conference, move it to Travel, and switch it to the model with the long
context.''} A milder form already exists: recommend a persona, then let the
user start it with one click. No single screen supports the full sequence. A
meta layer can, provided that each step is grounded, permissioned, and visible.
Composition is where chat stops being a help window and becomes an interface in
its own right.

\subsection{Accessibility and inclusion}

Natural-language access can lower barriers for users who find hierarchical
menus, small controls, or dense settings panels difficult. It can also help
users who are not fluent in the interface's language, since the assistant can
answer in the user's language and translate intent into the correct controls.

The deeper accessibility argument is that meta access provides a readable
interface to application state. Screen readers such as
JAWS~\cite{jaws2026} already do this
for graphical interfaces: they consume accessibility APIs and present a
structured, navigable description of widgets and content to blind and
low-vision users.
Model-based approaches to LLM-driven accessible interfaces pursue a related
goal \cite{jerry2026accessible}. Meta access extends the same principle from
widget-level descriptions to semantic application state and operations: the
assistant can say which conversations and topics exist, read current settings,
explain a feature, and offer a grounded next step. For a blind user, this is a
usable description of the workspace; for a user with a motor impairment, it
replaces precise pointing with intent; for a user with a cognitive or language
barrier, it can simplify navigation and explain controls in plain language.

The prototype already contains the ingredients: tutorials provide readable,
step-by-step guidance; metadata and links support orientation; preference
readouts avoid hunting through panels; personas lower configuration burden; and
confirmed proposals make changes inspectable before they are applied. These
benefits are not automatic. Meta access must not replace accessibility
compliance, keyboard navigation, or proper accessibility APIs, and it should
respect the same permissions and expose the same structure that a screen reader
sees rather than bypassing it. Latency, error recovery, and the risk of a
confident but wrong answer matter more, not less, when the assistant is the
user's primary route into the interface; the textual interface must remain fully
usable on its own.

\subsection{From chat to AI-era interfaces}

The chat client is a convenient first demonstrator, not the limit of the
pattern. An IDE can expose its project model, run configurations, and
diagnostics; a spreadsheet can expose sheets, ranges, formulas, and chart
definitions; a dashboard can expose widgets, filters, and data sources; an
operating system can expose files, applications, and settings. These are the
applications' existing internal models, not a new parallel representation; the
meta layer exposes them through a common contract. In each case the same three
ingredients recur: a workspace model that the assistant can inspect,
a capability layer that declares what may be read or changed, and evidence and
consent contracts that keep the assistant grounded and the user in control. The
chat-specific parts of the prototype, i.e., tutorials, personas, and
conversation links, are instances of these ingredients, not the ingredients
themselves. We
therefore propose that AI-era interfaces should expose such a meta layer as a
default part of their architecture, in the same way that they expose an API, a
settings model, or an accessibility layer.

\section{Designing for Whole-Interface Access}

\subsection{A portable meta-access contract}

The design should not depend on the chat surface. A portable meta layer needs
four pieces: a workspace model, ideally derived from the application's own
metamodel; a typed capability registry; an evidence and consent contract; and an
evaluation surface. Applications can implement these
pieces with different data models, transports, and vocabularies; the assistant
should see the same shape. Standardising the contract would let one assistant
work across applications without bespoke integrations, and would let users
compare and audit behaviour. The prototype realizes these pieces for a chat
client; the same pieces apply to other applications.

\subsection{Reuse application metamodels}

An assistant that works across arbitrary applications needs a model of each
application, not only a screenshot or a prompt. Most applications already have
one. A debugger is the clearest precedent: it exposes internal objects, call
stacks, and controlled mutations through a typed interface, without pretending
that the program is a flat text buffer. Meta access can generalize the same
idea. Instead of inventing a parallel description of the interface, the meta
layer should derive its workspace model and capability registry from the
application's existing metamodel---its document model, project model, schema,
scene graph, or component tree---and expose a stable, permissioned subset to the
assistant. The hard problem is not inventing a metamodel but choosing a minimal,
versioned, permissioned view and mapping it consistently across applications.

A general-purpose agentic harness has no single application metamodel to reuse.
Its workspace model is task-relative, and its stable part is the capability
registry and the policy that governs it. For such systems, the portable contract
should expose three layers instead: the task context (repository, files, and
running processes), the available capabilities and their scopes, and the harness
configuration that determines future capabilities. The first layer is
application-like, the second is the capability model described below, and the
third is a control plane: it should be readable and proposable, but not
self-applied when it touches authority.

\subsection{Model the interface as capabilities}

An action-capable meta layer needs an explicit capability model rather than a
giant prompt. Each capability should declare its inputs, its target object type,
its preconditions, whether it is read-only or mutating, whether it is
reversible, and what confirmation it requires. The model's job is to select and
fill capabilities, not to manipulate the DOM. A typed capability layer keeps
the assistant honest and makes the same capabilities available to buttons,
keyboard shortcuts, and the chat. This follows the tool-use pattern demonstrated
by API-calling agents \cite{patil2024gorilla,xi2023rise}. The persona catalogue
illustrates the split: reading persona metadata is a retrieval capability,
whereas instantiating a persona is a separate, mutating capability that the
user confirms by clicking a link.

\subsection{Guardrails as evidence and authority contracts}
\label{sec:guardrails}

The guardrails we propose are not content filters. They constrain what the meta
layer may claim, cite, disclose, and change. Each capability should declare a
contract with six parts: its evidence source, its permitted scope, how its
target is validated, what it discloses, what confirmation it requires, and what
it does when the contract cannot be satisfied. The contract is enforced by the
client and the capability layer, not only by the system prompt. A prompt can
ask the model not to invent an identifier; only code can reject a missing ID, a
stale deep link, an out-of-scope chat, or an unconfirmed mutation.

Table~\ref{tab:guardrails} maps these contracts onto the main capability
families. For message-level search, the critical boundary is evidence: the
answer must distinguish a summary from a message span, cite exact messages, and
respect the same permissions as the workspace. For navigation and tutoring, the
critical boundary is the target: the assistant must resolve the live interface
state, validate routes, highlight real controls, and explain unavailability
instead of inventing a setting. Persona recommendation adds a supply-chain
boundary: only catalogue metadata enters the meta context, and the definition
stays untrusted until the user starts it. Preference reads should be timestamped
snapshots of actual values, and mutations should require preview, confirmation,
and undo.

\begin{table}[htbp]
\centering
\small
\renewcommand{\arraystretch}{1.2}
\setlength{\tabcolsep}{4pt}
\begin{tabular}{@{}>{\raggedright\arraybackslash}p{0.16\linewidth}>{\raggedright\arraybackslash}p{0.30\linewidth}>{\raggedright\arraybackslash}p{0.48\linewidth}@{}}
\toprule
\textbf{Area} & \textbf{Example meta features} & \textbf{Guardrails} \\
\midrule
Message-level search &
Find where a decision was made; compare conversations; extract decisions or action items &
Cite exact messages and snippets; label summary versus full-text evidence; enforce workspace scope and permissions; mark extraction as inference; refuse comparisons when access is unclear \\
Navigation and tutoring &
Open a setting; deep-link to a chat, message, or topic; highlight UI elements; explain unavailable features &
Resolve live UI state and the capability registry; validate every deep link; highlight only real controls; state the reason for unavailability; never invent settings or routes \\
Persona recommendation and handoff &
Recommend a persona; explain categories; start a persona chat &
Expose catalogue metadata only; show the source; treat the definition as untrusted; import only on a user click; validate the model and settings before applying \\
Preference reading &
``What are my memory settings?'' &
Read current local values; timestamp the snapshot; distinguish a value from its control; explain the implications of the value \\
Mutating actions &
Rename or move a chat; change a setting; delete data &
Use a typed capability; show a preview or diff; require explicit confirmation; support undo; keep an audit log; grant least authority \\
\bottomrule
\end{tabular}
\caption{Guardrail contracts for meta-access capabilities. The same contract
applies at the point where the capability is exposed to the model and at the
point where the client executes or renders it.}
\label{tab:guardrails}
\end{table}

The contracts scale with the ladder. Levels 1--2 need evidence and target
validation. Level 3 adds provenance and scope controls over message content.
Levels 4--5 add preview, confirmation, reversibility, and audit. A multi-step
plan at level 5 should be checked against the contracts at every step, not only
at the end, and a failed contract should produce an explicit refusal or a
request for clarification rather than a confident fallback.

\subsection{Retrieval hierarchy and freshness}

The context strategy should be tiered. Level 2 uses the metadata catalogue,
the persona catalogue, and setting and preference values. Level 3 adds
summaries and a searchable index over message text and attachments, with
metadata as a pre-filter. This mirrors the retrieve-then-generate pattern of
retrieval-augmented generation \cite{lewis2020rag,gao2023rag}. Freshness should
be visible: the assistant should know when a summary was generated and should
prefer message-level evidence when the summary is old or the question is
specific. Size bounds should be explained rather than silent; if a query needs
material outside the assembled view, the assistant should say so and retrieve
further.

\subsection{Grounding, citation, and uncertainty}

Every factual claim about the workspace should be traceable. Links to
conversations are the minimal form of citation. Persona recommendations should
name only catalogue entries and link to their definitions so the user can
inspect the source before starting. Message-level retrieval should cite message
identifiers or snippets. Preference answers should name the settings section
and, where possible, link to it so the user can verify the value. When evidence
is missing, the assistant should say so and offer a next step (search full
text, broaden the time range, or ask the user). Calibrated
uncertainty is more useful than confident fabrication, especially when the
user is about to act on the answer \cite{amershi2019guidelines}.

\subsection{Confirmation, permissions, and reversibility}

Actions should follow a principle of least authority. Reading metadata,
reading message content, reading preference values, and mutating state are
distinct permissions. Bulk actions should require stronger confirmation than
single-object actions. Destructive actions should be reversible or explicitly
confirmed. The assistant should show a preview of the change and a diff where
applicable, and the user should be able to undo. These are standard
software-safety principles applied to a new kind of actor
\cite{horvitz1999principles,amershi2019guidelines}.

\subsection{Prompt injection and untrusted content}

A meta assistant reads content the user did not write, including model output,
imported chats, and document text. That content is untrusted and may contain
instructions aimed at the assistant. If the assistant can act, injected text
could try to trigger actions. Defences include treating retrieved content as
data rather than instructions, separating the instruction channel from the
content channel, restricting tools by scope, requiring confirmation for
mutations, and never allowing retrieved text to expand the assistant's
permissions \cite{perez2022ignore,greshake2023not,debenedetti2024agentdojo}.
Personas add a second boundary: their definitions are third-party content and
their system prompts can contain instructions. Keeping those prompts out of the
meta assistant's context, and importing them only in a separate user-started
chat, prevents a persona from steering the meta layer that recommends it.

\section{Risks and Open Problems}

\paragraph{Hallucinated workspace facts.}
A model that is asked about personal data will sometimes confabulate a title or
a topic. Grounding constraints and citations reduce but do not eliminate this.
Evaluation must measure precision of references, not only fluency.

\paragraph{Summary loss and staleness.}
Summaries are compressed, generated by a model, and may lag behind edits. The
prototype already acknowledges this in its prompt. A production system needs
freshness signals, backfills, and a retrieval path that can bypass summaries
when precision matters.

\paragraph{Context and cost limits.}
Size bounds on the assembled context are necessary but introduce silent blind
spots. Users need to know when material has been omitted, and the system should
degrade gracefully (for example by saying that some older material is outside
the current view).

\paragraph{Privacy and data minimisation.}
Sending metadata, settings, and preference values to a model endpoint is a real
disclosure.
Local-first designs can keep more processing on the client, and deployments
should make the boundary explicit. Memories deserve particular care because
they are curated personal facts rather than incidental text. Persona titles and
descriptions also leave the machine, and starting a persona fetches its
definition from an external repository, so the interface should distinguish
local data from catalogue traffic.

\paragraph{Persona catalogue trust.}
A shared persona repository is convenient but introduces supply-chain risk.
Titles and descriptions can be stale, duplicated, or deliberately misleading;
a definition can specify an unexpected model or sampling parameters, and its
system prompt may contain instructions that conflict with the host application.
The meta assistant should recommend only catalogue metadata, the start action
should show the persona's source and configuration, and imported definitions
should be treated as untrusted content. Ranking and freshness signals are part
of the retrieval problem, not just a presentation detail.

\paragraph{Standardisation and cross-application scope.}
If every application invents its own meta layer, assistants will remain bespoke
integrations. A common capability description, permission model, and audit
format would make meta access portable, but it also raises cross-application
questions: what happens when one request spans an IDE, a browser, and a chat
client, and which application owns consent? These are open design and governance
problems, not only engineering details.

\paragraph{Metamodel exposure and coupling.}
Reusing internal application models exposes details that may be unstable,
implementation-specific, or sensitive. A debugger-like interface must define
which internals are stable enough to expose, how to version them, and how to
avoid leaking secrets or bypassing application permissions. Coupling the
assistant to internal models also creates maintenance work as applications
evolve.

\paragraph{Reflexive configuration and self-escalation.}
A general-purpose harness such as OpenCode \cite{opencode2026} can change its
own model, tools, prompts, permissions, and hooks. If it can do so through a
generic file or shell tool, it can bypass a typed setting capability entirely.
Changes to authority or capability can persist across sessions, alter what the
assistant is allowed to do, and propagate to subagents; confirmation by the same
assistant is not a meaningful control. The policy layer must sit outside the
assistant's writable scope, changes that touch authority must be approved out of
band and audited, and activation should be explicit, preferably with a restart
and a rollback path.

\paragraph{Automation bias and loss of overview.}
If the chat becomes the primary entry point, users may stop learning the
interface and may over-trust its answers. The textual interface must remain
first-class, the assistant must be interruptible, and its actions must be
visible.

\paragraph{Evaluation.}
There is no standard benchmark for meta access. We need task suites built from
realistic workspaces, with known ground truth for retrieval and navigation, and
user studies that measure time, success, and trust. Safety evaluation must
include adversarial titles, injected summaries, and ambiguous requests.

\section{An Evaluation Agenda}

We propose evaluating meta access along five axes: retrieval quality, task success, grounding and calibration, safety and trust, and guardrail effectiveness.

\paragraph{Retrieval quality.}
Given a natural-language request, does the system return the correct
conversation or message, and does it avoid false positives? Useful metrics
include recall at small $k$, mean reciprocal rank, and reference precision
(the fraction of cited identifiers that exist and are relevant). For personas,
measure recommendation precision (did the suggested persona fit the stated
goal?) and handoff correctness (does the link start the definition it names?).

\paragraph{Task success.}
Can the user complete a real task---find a chat, understand a topic, read a
preference, find a suitable persona, change a setting---with fewer steps and
less time than the baseline interface? Measures include task completion, time
on task, number of turns, and abandonment.

\paragraph{Grounding and calibration.}
When the answer is wrong, does the system say so? We can measure the agreement
between stated confidence and correctness, and the rate of fabricated
references under missing or stale data. For preference questions, measure
whether the printed values match the client's current settings when the answer
is produced. Context construction is part of that outcome: Bousetouane
\cite{bousetouane2026context} argues that agents often fail through context
failure and proposes criteria for guardrail coverage, instruction consistency,
tool-schema quality, grounding sufficiency, and injection hardening. These
criteria apply directly to the bounded catalogue and excerpt pipeline of the
prototype.

\paragraph{Safety and trust.}
For action-capable features, measure the rate of unintended actions, the
effectiveness of confirmation, recovery after undo, and susceptibility to
prompt injection from retrieved content. For read-only features, measure
privacy leakage and user comfort with metadata, message excerpts, and setting
and preference disclosure.
Persona studies should include adversarial catalogue descriptions and
definitions, and check that the meta assistant never receives or repeats the
persona's system prompt.

\paragraph{Guardrail effectiveness.}
Each contract in Table~\ref{tab:guardrails} should be evaluated as a mechanism,
not only as a prompt instruction. Measure the unsupported-claim rate, citation
correctness, stale-link rate, scope violations, and disclosure accuracy for
read-only features. For navigation, measure whether deep links resolve to the
intended target and whether unavailable controls are explained correctly. For
mutations, measure confirmation compliance, preview accuracy, undo success, and
the rate of actions that violate declared permissions. Adversarial tests should
attempt to bypass the client-side checks, not only the system prompt.

\paragraph{Reflexive configuration.}
For harnesses, measure whether a proposed self-change is scoped, previewed, and
reversible; whether it persists across sessions; whether it expands tools,
permissions, or approval scope; and whether it propagates to subagents. Test
bypasses through generic file and shell tools, not only through the typed
setting capability. A useful benchmark is a set of configuration changes whose
effects on future capability are known, including changes that look benign in a
diff but alter authority after a restart.

Ablations should compare summaries only, full-text retrieval, and hybrid
retrieval; different catalogue sizes and recency caps; fresh versus stale
summaries and personas; and small versus large models. Longitudinal studies are
important because the value of meta access depends on the size and age of a
user's workspace.

\section{Conclusion}

The \metachat{} prototype shows that a chat assistant can already serve as a
useful access layer over a chat application: it can explain features, run
tutorials, retrieve conversations from metadata, search message text with
citations, recommend personas, read current preference values and per-chat
settings, ground its answers in links to chats, messages, topics, settings, and
personas, and guide users back into the interface. It also performs confirmed,
reversible changes to chats, settings, and memories. That covers levels 1--3
and a substantial part of level 4 of the ladder. The prototype already turns
the chat from a window onto a workspace into a natural-language entry point to
the application; full level-5 orchestration would extend that entry point from
confirmed single steps to composed, reviewable workflows.

Chat is not the only interface for LLMs, and general-purpose agentic harnesses
such as OpenCode \cite{opencode2026} are the right choice for many tasks that
operate on repositories, shells, or other external environments. Such a harness
is not an application with a fixed purpose: its workspace is the task, and its
own configuration is a control plane. If it can change that configuration, meta
access becomes reflexive, so authority changes need a separate out-of-band
boundary rather than a level-4S confirmation by the same assistant. But the chat
client remains one durable home for everyday conversation and for the personal
workspace that accumulates around it. Meta access is what lets the assistant
serve that workspace rather than only the current dialogue.

The opportunity is substantial: conversational search, navigation, onboarding,
organisation, and composition in one place. So is the responsibility. The same
properties that make meta access powerful---personal data, compressed context,
model-generated summaries, and the ability to act---make grounding, privacy,
confirmation, and reversibility essential design requirements rather than
polish. Guardrails turn those requirements into explicit evidence and authority
contracts that the client can enforce and evaluation can test. That is what lets
meta access turn the chat interface from a window onto a single conversation into
a conversational control layer for the whole workspace, extending chat toward
agentic capability without giving up explicit user control. We hope this draft
contributes a useful vocabulary and an evaluation agenda for building that
future carefully.

The broader proposal is that meta access should become a default property of
AI-era interfaces, not an optional feature of chat clients. The chat prototype
shows that the pattern is practical: a workspace model, grounded links, a
capability layer, and consent-gated mutations can coexist in a real application.
Accessibility is a direct beneficiary: a readable, permissioned model of an
application can complement screen readers such as
JAWS~\cite{jaws2026} and reduce
reliance on precise pointing, visual search, or memorised navigation. The next step is to
turn each application's existing metamodel into a stable,
permissioned meta-access interface, so the same contract works across IDEs,
dashboards, office suites, and operating systems, and to evaluate it as an
interface property rather than a single product feature.

\section*{Acknowledgements}

DeepSeek was used as a writing assistant during the preparation of this paper.
All AI-generated output was manually inspected, corrected, and revised by the
author(s), who remain responsible for the content and any errors.

\end{document}